\documentclass[runningheads]{llncs}
\usepackage[T1]{fontenc}
\usepackage{graphicx}
\usepackage{placeins}
\usepackage{float}
\usepackage{tabularx}
\usepackage{multirow} 
\begin{document}
\title{Pharmacokinetic State Space Models for Unbiased Prediction of Haemodynamic Collapse}
\titlerunning{PK-SSMs for Unbiased Prediction of Haemodynamic Collapse}
%
\author{Rithin Nagaraj\inst{1}\orcidID{0009-0008-9919-2466} \and
Sudiksha Chindula\inst{1}\orcidID{0009-0000-6396-4585} \and
Bhaskarjyoti Das\inst{2}\orcidID{0000-0003-1225-9354}}

\authorrunning{R. Nagaraj et al.}

\institute{Department of Computer Science and Engineering, PES University, Bengaluru, 560085, India\\
\email{rithin.nagaraj@gmail.com, sudiksha.chindula7@gmail.com} \and
Department of Computer Science and Engineering (AIML), PES University, Bengaluru, 560085, India\\
\email{bhaskarjyoti01@gmail.com}}
\maketitle              
\begin{abstract}
An Intraoperative Hypotension (IOH) event is a frequent complication during administration of general anaesthesia with serious downstream consequences, yet clinical management remains reactive and not predictive. Existing predictive models, however, ignore drug infusion history as a valuable signal for prediction despite its direct pharmacological relevance. Our model achieves an Area Under the Receiver Operating Characteristic curve (AUROC) of 0.7360 and an Area Under the Precision-Recall Curve (AUPRC) of 0.1794, representing a 2.73-fold lift over the random guessing AUPRC baseline (0.0657), with the removal of propofol and remifentanil effect-site concentrations resulting in a 13.9\% AUPRC drop compared to the full model. This is consistent with the hypothesis that pharmacokinetic trajectories encode impending haemodynamic changes before they manifest in the Mean Arterial Pressure (MAP). Additionally, this paper shows that training without lead-gap filtering degraded AUROC by 16.7\%, empirically confirming that unfiltered models learn to detect ongoing hypotension rather than predict future events. Finally, a Mamba-based architecture achieves the aforementioned high prediction performance while maintaining a constant memory footprint across a range of sequence lengths, unlike the quadratic VRAM overhead typical of vanilla Transformers, making it the more practical choice for continuous intraoperative deployment.

\end{abstract}

\section{Introduction}
General anaesthesia for surgical interventions routinely involves administrating hypnotics and opioid analgesics to induce the loss of consciousness and tolerance to surgery.~\cite{ref1} Commonly used anaesthetics interfere with the cardiovascular system by reducing cardiac inotropy and systemic vascular resistance, ultimately leading to hypotension.~\cite{ref2}

Intraoperative hypotension (IOH), conventionally defined as a mean arterial pressure below 65 mmHg,~\cite{ref3} sustained for one or more consecutive minutes, is among the most clinically consequential adverse events encountered during general anaesthesia. An IOH event is potentially harmful, being linked to conditions such as myocardial injury~\cite{ref4}, kidney injury~\cite{ref4,ref5}, delirium~\cite{ref6} and post-operative nausea and vomiting.~\cite{ref7} The standard clinical response is reactive wherein anaesthesiologists administer vasopressors or fluids after a hypotensive event has already been detected. However, the actual prevention of the episode entirely would require accurate prediction of hypotension in advance.

This imperative has motivated a sustained effort to develop predictive models for IOH. However, these approaches share three critical limitations: first, they overlook drug infusion history as a pharmacologically informative precursor to an IOH event; second, they predominantly employ transformer architectures whose self-attention mechanism scales with $O(N^2)$ complexity, rendering them impractical for real-time intraoperative deployment on edge devices; and third, they suffer from lead gap contamination i.e., training on windows containing the onset of an IOH event, thereby conflating genuine forecasting with retrospective detection of deterioration already underway. The contributions in this paper are as follows:
\begin{itemize}
  \item The central claim is that incorporating remifentanil and propofol effect-site concentration (CE) trajectories as explicit model inputs produces a measurable, independent predictive contribution beyond haemodynamic signals alone.
  \item A secondary but practically significant contribution of this work concerns the architectural choice. We propose replacing the transformer backbone with a Mamba selective state space model~\cite{ref8} whose recurrent inference mode operates in $O(1)$ per-timestep VRAM and enables streaming computation.
  \item Lastly, we experimentally show how training on datasets with lead gap contamination produces models with inflated internal performance and near-zero clinical specificity when evaluated on filtered held-out data. We are also able to quantify the impact of this contamination on reported Area Under Receiver Operating Characteristic Curve (AUROC).
\end{itemize}

\section{Related Work}\label{sec2}

\subsection{IOH Prediction with Deep Learning}\label{sec2subsec1}
The prediction of intraoperative hypotension has progressed from classical feature-engineering to end-to-end deep learning on raw biosignals. The Hypotension Prediction Index (HPI) by Hatib et al.~\cite{ref9} established the paradigm, using arterial waveform features as inputs to a logistic regression model and achieving an AUROC of 0.92 for binary prediction up to 15 minutes in advance. Lee et al.~\cite{ref11} demonstrated that deep learning on raw arterial pressure waveforms from VitalDB outperforms classical regression, establishing VitalDB as the standard benchmark for this task. The current state of the art is represented by Shim et al.~\cite{ref12}, who combine four biosignal waveforms (ABP, ECG, PPG, ETCO2) with preoperative covariates in a hybrid CNN-RNN architecture, achieving an AUROC of 0.94 on 2611 VitalDB patients which is the primary academic benchmark against which we position our work. Kapral et al.~\cite{ref1} are notable for incorporating intraoperative medication data (including propofol) as time-stamped inputs to a Temporal Fusion Transformer forecasting continuous MAP trajectories. However, they did not compute effect-site concentrations and thus captured only the dosing event rather than the pharmacokinetic trajectory.

\subsection{Drug Pharmacokinetics and Haemodynamic Consequence}\label{sec2subsec2}
Propofol exerts cardiovascular effects through peripheral vasodilation and myocardial depression, reducing MAP in a manner governed by blood-to-effect-site equilibration kinetics.~\cite{ref13,ref14} In the Schnider PK model, the $k_{e0}$ for propofol implies a blood-to-effect-site half-life of roughly 1.5 minutes,~\cite{ref15,ref16} meaning peak haemodynamic consequence follows peak plasma concentration by a clinically meaningful interval. A similar lag exists for remifentanil under the Minto model.~\cite{ref17} Critically, a model with access to the current effect-site concentration trajectory can estimate committed cardiovascular perturbation before it manifests haemodynamically. Despite this well-established basis, no prior work has utilized CE trajectories from intraoperative dosing records as explicit predictive features. Our work is the first, to our knowledge, to do so for both propofol and remifentanil using the VitalDB dosing record.

\subsection{Sequence Modelling Architectures for Clinical Time Series}\label{sec2subsec3}
LSTMs and GRUs remain competitive for clinical time series prediction,~\cite{ref18} while Transformer architectures have demonstrated strong results including continuous MAP forecasting.~\cite{ref1} However, the quadratic scaling of self-attention with sequence length constrains real-time deployment: continuous streaming inference requires recomputing the full attention matrix at each timestep, incompatible with low-latency operation without dedicated GPU hardware. State space models (SSMs) address this with $O(1)$ per-timestep recurrence. The Mamba architecture~\cite{ref8} extends classical SSMs with an input-dependent selection mechanism that recovers much of the Transformer's selective attention while retaining linear scaling.

\subsection{Dataset Methodology in IOH Prediction}\label{sec2subsec4}
While recent predictive models report high discrimination capabilities, their underlying dataset extraction methodologies have come under increasing scrutiny. Enevoldsen and Vistisen~\cite{ref10} demonstrated that HPI~\cite{ref9} and subsequent validation studies~\cite{ref19} evaluated performance using observation windows not strictly separated from hypotensive events by a designated blind spot interval (lead-gap), leading to performance overestimation due to selection bias. Specifically, when a model observes a window where haemodynamic decline toward 65 mmHg has already begun, the task shifts from forecasting de novo hypotension to detecting an ongoing deterioration, a continuation bias theoretically identified as inflating reported performance. Prior to this work, the quantitative impact of lead-gap contamination on reported AUROC metrics across large-scale IOH datasets remained unresolved.

\section{Methodology}\label{sec3}
\begin{figure}[h!]
    \centering
    \includegraphics[width=0.75\linewidth]{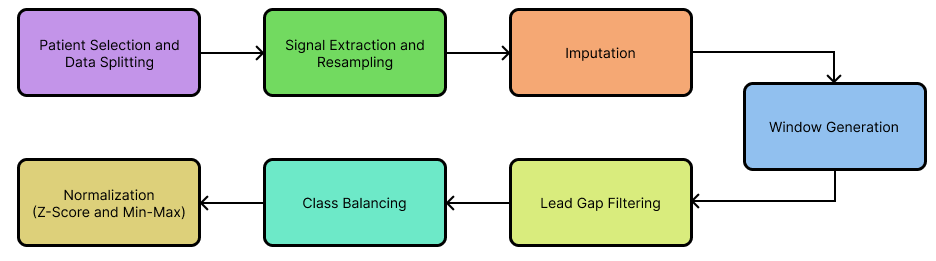}
    \caption{The data pre-processing pipeline, comprising five sequential stages: signal extraction and resampling, gap handling and imputation, sliding-window generation with lead-gap filtering, class balancing, and normalization.}
    \label{fig:data_pipeline}
\end{figure}

The data preprocessing pipeline consisted of five sequential stages: signal extraction and resampling, gap handling and imputation, sliding-window generation with lead-gap filtering, class balancing, and normalization (Figure~\ref{fig:data_pipeline}).

\subsection{Dataset and Patient Population}\label{sec3subsec1}
The VitalDB~\cite{ref20} open dataset contains high-fidelity, intraoperative biosignals and clinical metadata from 6,388 surgical cases collected at the Seoul National University Hospital between June 2016 and August 2017. Cases were included if all four required sequential signal tracks---ART\_MBP (MAP), HR (Heart Rate), PPF20\_CE (Propofol Effect-Site Concentration) and RFTN20\_CE (Remifentanil Effect-Site Concentration)---were recorded continuously throughout the procedure. After applying inclusion criteria and lead-gap filtering, 3,419 patients were retained (2,721 training, 698 test). Lead-gap filtering is where any window in which MAP fell below 65 mmHg at any point during the lead-gap interval was discarded. Patients were randomly assigned to training and test sets at an 80:20 ratio at the patient level (random seed 42), prior to any windowing or normalization, to prevent data leakage.

\FloatBarrier

\vspace{-6mm}

\subsection{Signal Extraction and Preprocessing}\label{sec3subsec2}
The extracted data was resampled to 0.5Hz and drug concentration values with missing entries were imputed to zero, reflecting the absence of active drug infusion. For haemodynamic signals, a three-tier gap-handling strategy was applied: gaps under one minute were filled via linear interpolation; gaps between one and five minutes were filled via forward-fill method; gaps exceeding five minutes were flagged and all overlapping observation windows were discarded. Static demographic features were extracted from the VitalDB clinical metadata table for each case, comprising age, sex, height, weight, and BMI. Missing demographic values were imputed using the median of each feature computed exclusively from the training cohort.

\subsection{Window Generation and Labelling}\label{sec3subsec3}
The windows generated consisted of three sub-windows: a 30-minute observation window (900 samples), a 5-minute lead gap window (150 samples), and a 10-minute prediction window (300 samples) at 0.5 Hz sampling frequency. Windows were extracted using a sliding step of 30 samples (1 min). A window was assigned a target label of 1 if MAP fell strictly below 65 mmHg for at least 30 consecutive samples (1 min) within the prediction window~\cite{ref3} and windows not meeting this criterion were assigned 0 as the label.

\subsection{Dataset Variants}\label{sec3subsec4}

Multiple Dataset Variants, where D1 is the full dataset and D2–D5 are
ablations, were used to isolate the contribution of individual signal groups.

\begin{table}[htbp]
\centering
\caption{Dataset summary and ablation configurations.}
\label{tab:dataset_summary}
\setlength{\tabcolsep}{8pt}
\renewcommand{\arraystretch}{1.3}
\begin{tabularx}{\textwidth}{l >{\raggedright\arraybackslash}X >{\raggedright\arraybackslash}X}
\hline
\textbf{Dataset} & \textbf{Signals} & \textbf{Notes} \\
\hline
D1 & ART\_MBP, HR, PPF20\_CE, RFTN20\_CE + patient demographics & Full Dataset \\
D2 & ART\_MBP, HR + patient demographics & Drug Ablation \\
D3 & Same as D1 with no lead-gap filtering & Lead-gap filter disabled; all previously discarded windows due to lead-gap filter retained. \\
D4 & Same as D1 with no patient demographics & Demographic Ablation \\
D5 & PPF20\_CE, RFTN20\_CE + patient demographics & Vitals Ablation \\
\hline
\end{tabularx}
\end{table}
\FloatBarrier

\vspace{-6mm}

\subsection{Normalization and Class Balancing}\label{sec3subsec5}
The sequential data was normalized by z-score normalization and the static demographic data were normalized by min-max normalization, both fitted exclusively on the training data and applied to the test data without refitting. The training set was undersampled to a 1:3 positive-to-negative split to mitigate the effect of class imbalance during training. The test set was intentionally kept unbalanced to preserve the natural prevalence of 6.6\% positive instances for unbiased evaluation.

\subsection{Model Architecture}\label{sec3subsec6}
We used a Mamba State Space Model which is a Selective State Space Model. These models show $O(N)$~\cite{ref8} time complexity with respect to N sequence length during training, in contrast to the $O(N^2)$ attention mechanism of Transformers, making them substantially more memory-efficient for long physiological sequences.

\begin{figure}[h!]
    \centering
    \includegraphics[width=1.0\linewidth]{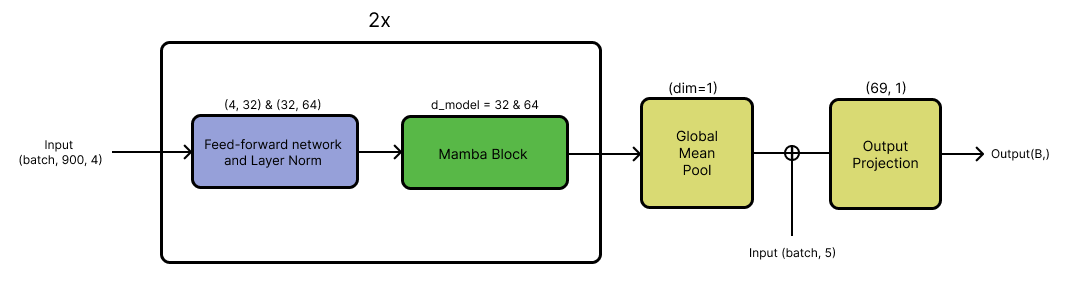}
    \caption{The Mamba-based model architecture. The d\_model varies per block.}
    \label{fig:model_architecture}
\end{figure}

The Mamba blocks have the same configuration (d\_state=16, d\_conv=4, expand=2) with changes in d\_model as shown in Figure~\ref{fig:model_architecture}. The output projection is a two-layer MLP with ELU activation to produce a scalar logit. Total trainable parameters: 47,297.

We also trained a Transformer model with the same two-block configuration and d\_model with a 1D-Convolutional layer (in\_channels=32, out\_channels=64, kernel\_size=5, stride=2) in-between the transformer blocks which reduced the sequence length from 900 to 448 timesteps, exploiting the fact that haemodynamic signals at 0.5 Hz evolve over minutes rather than seconds. Total trainable parameters: 75,425. The 1D-Convolutional layer was not applied to the Mamba model, which does not exhibit quadratic scaling with sequence length and therefore does not require it.

The Transformer was trained exclusively on Dataset D1 for architecture comparison purposes. All ablation experiments were conducted using the Mamba architecture due to its linear computational scaling, making it practical to train across multiple dataset variants.

\subsection{Training Procedure}\label{sec3subsec7}
All models were trained using AdamW optimizer with a learning rate of $1\times10^{-4}$ and a weight decay of $1\times10^{-4}$. Binary cross-entropy with logits (BCEWithLogitsLoss) was used as the training objective. A ReduceLROnPlateau scheduler with patience 3 and early stopping with patience 5 were applied on validation Area Under Precision-Recall Curve (AUPRC), restoring the best validation weights. Gradient clipping was applied with a maximum norm of 1.0 to stabilize the training. The batch size was 32 and the training ran for a maximum of 50 epochs. A 20\% split at the patient level of the training set was reserved as a validation set for scheduler monitoring, early stopping; this validation set is entirely separate from the held-out test set. Each model was trained independently on 3 random seeds (42, 123, 7). All experiments were conducted on an NVIDIA L4 GPU.

\subsection{Evaluation Metrics}\label{sec3subsec8}
AUPRC was selected as the primary metric given the class imbalance (6.6\% positive prevalence), with a random guessing baseline of 0.0657. AUROC was used as a secondary metric. AUPRC and AUROC were reported with a 95\% Confidence Interval via stratified bootstrap (1000 iterations). For statistical testing and Confidence Interval generation, the predicted probabilities from the three seeds were ensembled via an unweighted arithmetic mean prior to bootstrapping. A single operating threshold was selected by maximising Youden's J statistic $(sensitivity + specificity - 1)$ from the Mamba model trained on Dataset D1, on its validation set using seed 42, and applied uniformly across all models to enable consistent comparison of threshold-dependent metrics. At the specified threshold---Sensitivity, Specificity, Positive Predictive Value (PPV), Negative Predictive Value (NPV) and F1-score were calculated for each model. Two clinical utility metrics were additionally computed: mean lead time in minutes, defined as the time from the end of the observation window to the onset of the corresponding IOH event within the prediction window, and false alarm rate expressed as the number of false positive predictions per hour of surgery. DeLong test for AUROC differences between model pairs~\cite{ref27} was performed with Bonferroni correction for five simultaneous comparisons (corrected threshold $p < 0.01$).

\section{Results}\label{sec4}
\subsection{Dataset Characteristics}\label{sec4subsec1}

\begin{table}[htbp]
\centering
\caption{Dataset characteristics for train and test splits.}
\label{tab:dataset_characteristics}
\renewcommand{\arraystretch}{1.2}
\setlength{\tabcolsep}{4pt}
\small
\begin{tabularx}{\textwidth}{
p{3.8cm}
l
>{\raggedleft\arraybackslash}X
>{\raggedleft\arraybackslash}X
>{\raggedleft\arraybackslash}X
>{\raggedleft\arraybackslash}X
}
\hline
\textbf{Dataset} & \textbf{Split} & \textbf{Patients} & \textbf{Positive windows} & \textbf{Negative windows} & \textbf{Total windows} \\
\hline
\multirow{2}{3.8cm}{Filtered (with lead gap filtering)} & train & 2,721 & 24,004 & 72,012 & 96,016 \\
 & test & 698 & 6,096 & 86,655 & 92,751 \\
\hline
\multirow{2}{3.8cm}{Contaminated (without lead gap filtering)} & train & 2,747 & 78,602 & 235,806 & 314,408 \\
 & test & 701 & 19,394 & 103,963 & 123,357 \\
\hline
\end{tabularx}
\end{table}
\FloatBarrier

\begin{table}[htbp]
\centering
\caption{Patient demographics for train and test splits after lead-gap filtering and balancing. Values are mean $\pm$ standard deviation unless otherwise noted.}
\label{tab:demographics}
\renewcommand{\arraystretch}{1.3}
\begin{tabular}{l c c c}
\hline
\textbf{Split} & \textbf{Age} & \textbf{Sex (\% Female)} & \textbf{Mean BMI} \\
\hline
Train & 58.5 $\pm$ 15.2 & 44.4\% & 23.0 $\pm$ 3.6 \\
Test  & 58.3 $\pm$ 15.2 & 43.7\% & 23.1 $\pm$ 3.5 \\
\hline
\end{tabular}
\end{table}
\FloatBarrier

\noindent Total surgical duration of the test cohort was 1545.8 hours (698 patients); per-patient procedure duration was not separately recorded. The two splits are well-matched on all demographic covariates (Table~\ref{tab:demographics}).

\subsection{Primary Model Performance}\label{sec4subsec2}
\begin{figure}[h!]
    \centering
    \includegraphics[width=1.0\linewidth]{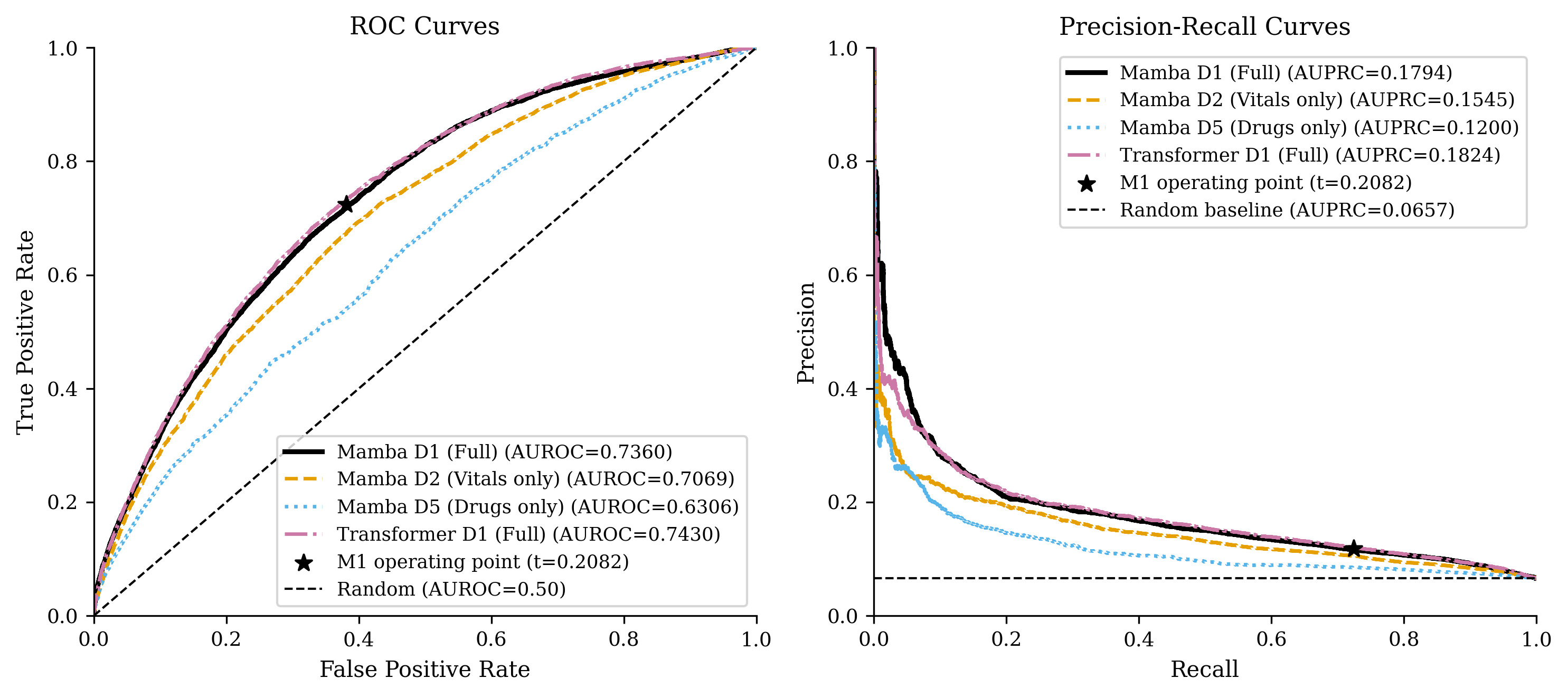}
    \caption{AUROC and AUPRC curves for all model variants.}
    \label{fig:auroc_auprc}
\end{figure}

We define M1--M5 as Mamba models trained on Datasets D1--D5 respectively, and T1 as the Transformer trained on D1. All results are evaluated on a fixed held-out test set of 698 patients and 92,751 windows at a natural positive prevalence of 6.6\%. ROC and precision-recall curves for all models are shown in Figure~\ref{fig:auroc_auprc}.

\begin{table}[htbp]
\centering
\caption{Model classification performance across different dataset configurations.}
\label{tab:model_classification}
\footnotesize
\renewcommand{\arraystretch}{1.2}
\setlength{\tabcolsep}{3pt}
\begin{tabularx}{\textwidth}{
l
>{\raggedright\arraybackslash}X
>{\raggedright\arraybackslash}X
c c c c c
}
\hline
\textbf{Model} & \textbf{AUROC} & \textbf{AUPRC} & \textbf{Sens} & \textbf{Spec} & \textbf{PPV} & \textbf{NPV} & \textbf{F1} \\
\hline
M1 & 0.7360 (0.730--0.742) & 0.1794 (0.172--0.189) & 0.724 & 0.619 & 0.118 & 0.970 & 0.203 \\
M2 & 0.7069 (0.700--0.713) & 0.1545 (0.147--0.162) & 0.747 & 0.553 & 0.105 & 0.969 & 0.184 \\
M3 & 0.6126 (0.605--0.620) & 0.1005 (0.096--0.105) & 0.986 & 0.044 & 0.068 & 0.978 & 0.127 \\
M4 & 0.7396 (0.734--0.746) & 0.1794 (0.172--0.188) & 0.720 & 0.637 & 0.123 & 0.970 & 0.209 \\
M5 & 0.6306 (0.624--0.638) & 0.1200 (0.114--0.127) & 0.795 & 0.371 & 0.082 & 0.963 & 0.148 \\
T1 & 0.7430 (0.737--0.749) & 0.1824 (0.175--0.191) & 0.797 & 0.549 & 0.111 & 0.975 & 0.194 \\
\hline
\end{tabularx}
\end{table}
\FloatBarrier

\noindent The operating threshold of 0.2082 was fixed across all models using Youden's J on the M1 validation set. False alarm rates at the operating threshold were 21.38, 25.03, 53.61, 20.33, and 35.28 alarms/hour for M1 through M5 respectively, and 25.30 alarms/hour for T1. The mean lead time in minutes was $9.926 \pm 2.473$ across all datasets.

\FloatBarrier

\begin{figure}[htbp]
    \centering
    \begin{minipage}{0.48\textwidth}
        \centering
        \includegraphics[width=\linewidth]{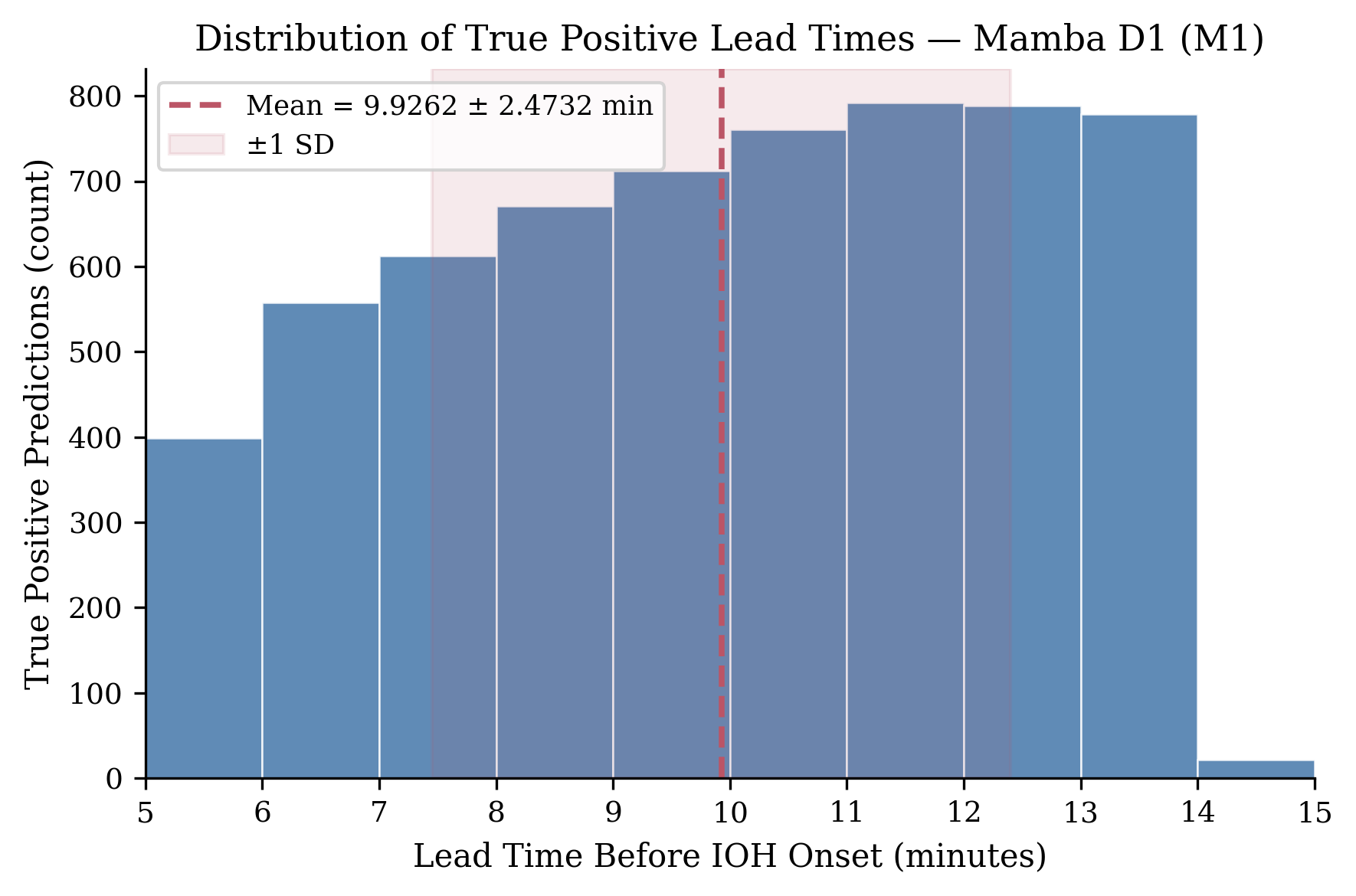}
        \caption{Distribution of mean lead time for true positive predictions (M1).}
        \label{fig:lead_time}
    \end{minipage}\hfill
    \begin{minipage}{0.48\textwidth}
        \centering
        \includegraphics[trim=0.25cm 0.25cm 0.25cm 0.25cm, clip, width=\linewidth]{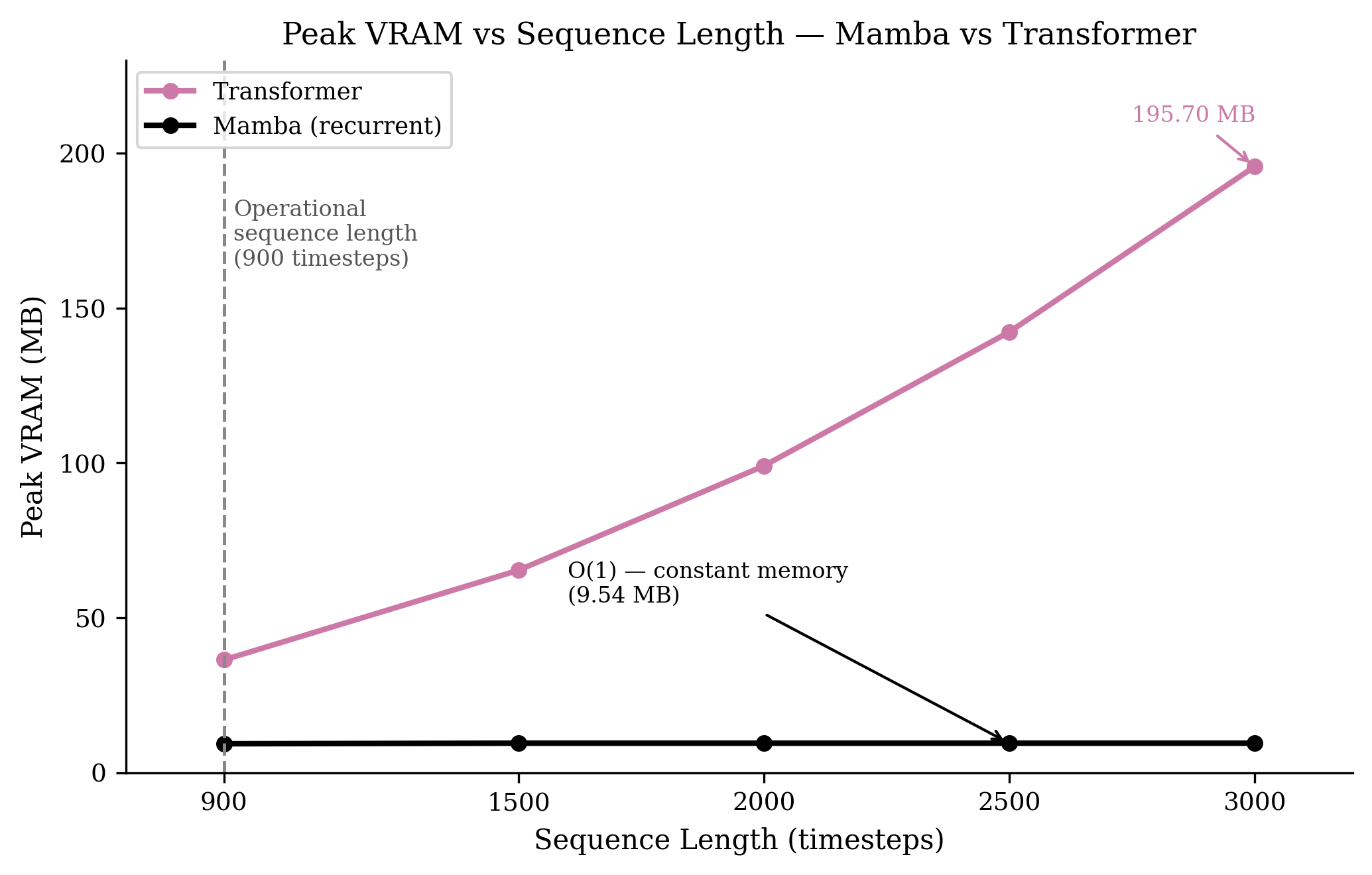}
        \caption{Peak VRAM (MB) vs.\ sequence length for Mamba and Transformer.}
        \label{fig:vram_efficiency}
    \end{minipage}
\end{figure}
\FloatBarrier

\subsection{Signal Ablation}\label{sec4subsec3}
A DeLong test with Bonferroni-corrected threshold of $p < 0.01$ ($\alpha=0.05$) was performed for five comparisons: M1 vs M2, M1 vs M3, M1 vs M4, M1 vs M5, and M1 vs T1.

\begin{table}[htbp]
\centering
\caption{Pairwise DeLong test for AUROC comparisons between models.}
\label{tab:statistical_comparison}
\renewcommand{\arraystretch}{1.3}
\begin{tabular}{l c c r c c}
\hline
\textbf{Models tested (A vs B)} & \textbf{AUROC A} & \textbf{AUROC B} & \textbf{$z$} & \textbf{$p$} & \textbf{$p_{\mathrm{bonferroni}}$} \\
\hline
M1 vs M2 & 0.7360 & 0.7069 & 13.197 & $<0.0001$ & $<0.0001$ \\
M1 vs M3 & 0.7360 & 0.6126 & 32.160 & $<0.0001$ & $<0.0001$ \\
M1 vs M4 & 0.7360 & 0.7396 & $-5.534$ & $<0.0001$ & $<0.0001$ \\
M1 vs M5 & 0.7360 & 0.6306 & 28.332 & $<0.0001$ & $<0.0001$ \\
M1 vs T1 & 0.7360 & 0.7430 & $-4.958$ & $<0.0001$ & $<0.0001$ \\
\hline
\end{tabular}
\end{table}
\FloatBarrier

\noindent Haemodynamic vitals carry the most valuable learning signal for the Mamba model. This is evident from the drop in AUROC when comparing M1 and M5 (Table~\ref{tab:statistical_comparison}, M1 vs M5). Drug history (PPF20\_CE and RFTN20\_CE) also carries a significant learning signal: removing it reduced AUROC by 3.95\% (Table~\ref{tab:statistical_comparison}, M1 vs M2). Static patient-specific demographics had minimal independent contribution, as shown by the negligible AUROC change between M1 and M4 (Table~\ref{tab:statistical_comparison}, M1 vs M4).

\subsection{Lead-Gap Contamination}\label{sec4subsec4}
Training on the contaminated dataset (D3, 314,408 training windows) without lead-gap filtering reduced AUROC from 0.7360 to 0.6126 relative to the filtered model (D1, 96,016 training windows), a 16.7\% relative degradation despite the contaminated set containing $3.3\times$ more training data. The DeLong test confirms this difference is statistically significant after Bonferroni correction (Table~\ref{tab:statistical_comparison}, M1 vs M3, $p<0.01$).

\subsection{Architecture Comparison}\label{sec4subsec5}

\begin{table}[htbp]
\centering
\caption{Performance and resource usage of Mamba vs. Transformer on Dataset D1.}
\label{tab:model_comparison}
\renewcommand{\arraystretch}{1.3}
\begin{tabular}{l c c}
\hline
\textbf{Metric} & \textbf{Mamba} & \textbf{Transformer} \\
\hline
Parameters & 47,297 & 75,425 \\
Inference latency & 1.726 ms & 1.511 ms \\
Peak VRAM at 900 timesteps & 9.34 MB & 36.42 MB \\
Peak VRAM at 3000 timesteps & 9.54 MB & 195.70 MB \\
Training time per epoch & 23.95 s & 117.08 s \\
AUROC & 0.7360 & 0.7430 \\
AUPRC & 0.1794 & 0.1824 \\
\hline
\end{tabular}
\end{table}
\FloatBarrier

\noindent Both models exhibit latency well within the 2000~ms real-time budget (Table~\ref{tab:model_comparison}). However, Mamba's peak VRAM usage remains essentially stable---increasing by only 0.20~MB when the sequence length scales from 900 to 3000 timesteps---whereas the Transformer's VRAM grows from 36.42~MB to 195.70~MB over the same range (Figure~\ref{fig:vram_efficiency}). Against a marginal AUROC difference of 0.007 in favour of the Transformer, Mamba's constant memory footprint supports deployment on memory-constrained clinical hardware independently of recording duration.

\section{Discussions}\label{sec5}
The results of the Mamba model (M1 in Table~\ref{tab:model_classification}) are contextualised against the commercial Hypotension Prediction Index (HPI), which reports AUROC of 0.95--0.97 at prediction horizons of 5 to 15 minutes,~\cite{ref21} and against a deep learning approach achieving AUROC of 0.9145 (internal) and 0.9035 (external validation) using 90-second arterial pressure waveforms sampled at 100~Hz for 10-minute IOH prediction without hand-engineered features.~\cite{ref22} The gap between our results and these benchmarks reflects three deliberate constraints rather than architectural limitations: our signals are downsampled to 0.5~Hz, discarding waveform morphology that high-frequency models exploit; our dataset is restricted to Total Intravenous Anaesthesia (TIVA) cases from a single-centre; and our lead-gap filter removes windows that contaminated prior work, producing a strictly harder prediction task. The PPV of 0.118 reflects the 6.6\% positive prevalence mathematically suppressing precision regardless of model quality; the commercially deployed HPI reports a comparable 0.126 PPV,~\cite{ref26} confirming this as a structural feature of low-prevalence IOH prediction rather than a deficiency of either system. Operating threshold can be adjusted to reduce false alarms at the cost of sensitivity depending on clinical deployment requirements.

The most clinically notable finding of this work is the independent predictive contribution of pharmacokinetic effect-site concentrations, reducing AUPRC by 13.9\% (0.1794 to 0.1545) and AUROC by 3.95\% on removing drug history from the full D1 signal set (Table~\ref{tab:statistical_comparison}, M1 vs M2). This is not a trivial finding: haemodynamic signals are concurrent correlates of cardiovascular state, whereas effect-site concentrations encode the pharmacokinetic trajectory of drugs already administered, providing information about impending haemodynamic changes before those changes are reflected in the MAP signal, a temporal advantage not available to models relying on haemodynamic signals alone. This finding is consistent with prior work demonstrating that visualising real-time drug effect-site concentrations improved clinical decision-making and haemodynamic control,~\cite{ref23,ref24} and extends that observation into the domain of automated early warning, suggesting that pharmacokinetic awareness should be a design requirement, not an optional feature, for IOH prediction systems deployed in TIVA cases.

Training without lead-gap filtering reduced AUROC despite the contaminated dataset containing substantially more training windows (314,408 vs 96,016; Table~\ref{tab:dataset_characteristics}). The 16.7\% AUROC degradation observed here provides empirical quantification of this effect. This finding is consistent with Enevoldsen \& Vistisen,~\cite{ref10} who demonstrated that the Hypotension Prediction Index's reported performance is likely overestimated due to an equivalent selection bias in its training methodology, and with the subsequent retraction of the primary conclusion of a major HPI validation study.~\cite{ref25} We therefore argue that lead-gap filtering should be adopted as a methodological standard for IOH prediction evaluation, and that published AUROC figures from work without equivalent filtering should be interpreted with caution.

While both architectures comfortably satisfy the real-time latency budget for 0.5 Hz physiological sampling, the critical deployment distinction lies in memory scaling. As shown in Table~\ref{tab:model_comparison} and Figure~\ref{fig:vram_efficiency}, Mamba maintains a constant VRAM footprint across extending sequence lengths, whereas the Transformer's memory cost grows quadratically. For continuous intraoperative monitoring that may span several hours, this scaling behavior is practically significant. Despite a marginal AUROC deficit, Mamba’s ability to operate strictly within the memory envelope of standard clinical edge hardware justifies it as the superior architecture for real-time IOH warning systems.

Several limitations constrain this work. First, our model was evaluated on a single-centre cohort restricted to TIVA, as pharmacokinetic trajectories for volatile anaesthetics differ fundamentally and are rarely recorded. Second, our deliberate downsampling to 0.5 Hz discards arterial waveform morphology, which high-frequency commercial systems exploit for superior discrimination, and restricts the input space to four core signals, leaving the predictive value of high-frequency ECG or capnography unexplored. Architecturally, Mamba’s recurrent inference requires a "warm-up" period; unlike window-based models that withhold outputs until the buffer is full, early streaming predictions may suffer from poor calibration before the hidden state accumulates sufficient temporal context. The duration and prediction impact of this warm-up remain unquantified. Finally, the false alarm rate of 21.38 per hour poses a significant alert fatigue risk; prospective trials should evaluate threshold optimisation and clinical integration before deployment.

\section{Conclusion}\label{sec6}
This work demonstrates that intraoperative hypotension can be predicted between 5 and 15 minutes in advance using a Mamba state space model trained on routinely available haemodynamic and pharmacokinetic signals, achieving an AUROC of 0.7360 and a 2.73-fold AUPRC lift over random guessing. We show empirically that the absence of lead-gap filtering degrades AUROC by 16.7\%, providing direct quantification of a contamination effect that likely affects published performance figures across prior IOH prediction literature. The independent predictive contribution of Propofol and Remifentanil effect-site concentrations suggests that pharmacokinetic awareness should be a design requirement for IOH prediction systems deployed in TIVA cases. Additionally, Mamba's constant memory footprint of $\sim$9.34~MB across all tested sequence lengths supports real-time deployment on standard clinical hardware without dedicated GPU infrastructure. Prospective multi-centre validation across anaesthetic regimens and operating room trials remain the critical next steps toward clinical translation.

\bibliographystyle{splncs04}
\bibliography{references}


\end{document}